\documentclass[runningheads]{llncs}

\usepackage{eccv}

\usepackage{eccvabbrv}

\usepackage{graphicx}
\usepackage{booktabs}
\usepackage{xcolor}
\usepackage{amsmath}
\usepackage{amssymb}
\usepackage{array}
\usepackage{multirow}
\usepackage{color}
\usepackage{colortbl}
\usepackage{framed}
\usepackage{bm}
\usepackage{bbm}
\usepackage{xspace}
\usepackage{enumitem}
\usepackage{xparse}
\usepackage{algorithm}
\usepackage{listings}
\usepackage{url}
\usepackage{wrapfig}

\usepackage[accsupp]{axessibility}  

\definecolor{Gray}{gray}{0.6}
\definecolor{TitleColor}{gray}{0.95}
\definecolor{LightCyan}{RGB}{255,228,234}
\definecolor{blond}{rgb}{0.98, 0.94, 0.75}

\def \ie {\emph{i.e.}}
\def \eg {\emph{e.g.}}
\def \etal {\emph{et al.}}

\newcommand{\tit}[1]{\smallbreak\noindent\textbf{#1.}}
\newcommand{\tinytit}[1]{\noindent\textbf{#1.}}

\newcommand{\ours}{BRIDGE\xspace}

\usepackage{hyperref}

\usepackage{orcidlink}

\begin{document}
\sloppy

\title{BRIDGE: Bridging Gaps in Image Captioning Evaluation with Stronger Visual Cues} 

\titlerunning{Bridging Gaps in Image Captioning Evaluation with Stronger Visual Cues}

\author{Sara Sarto\inst{1}\orcidlink{0000-0003-1057-3374} \and
Marcella Cornia\inst{1}\orcidlink{0000-0001-9640-9385} \and
Lorenzo Baraldi\inst{1}\orcidlink{0000-0001-5125-4957} \and
Rita Cucchiara\inst{1,2}\orcidlink{0000-0002-2239-283X}}

\authorrunning{S.~Sarto et al.}

\institute{University of Modena and Reggio Emilia, Italy \\
\and
IIT-CNR, Italy\\
\email{\{name.surname\}@unimore.it}}

\maketitle

\begin{abstract}
Effectively aligning with human judgment when evaluating machine-generated image captions represents a complex yet intriguing challenge. Existing evaluation metrics like CIDEr or CLIP-Score fall short in this regard as they do not take into account the corresponding image or lack the capability of encoding fine-grained details and penalizing hallucinations. To overcome these issues, in this paper, we propose \ours, a new learnable and reference-free image captioning metric that employs a novel module to map visual features into dense vectors and integrates them into multi-modal pseudo-captions which are built during the evaluation process. This approach results in a multimodal metric that properly incorporates information from the input image without relying on reference captions, bridging the gap between human judgment and machine-generated image captions. Experiments spanning several datasets demonstrate that our proposal achieves state-of-the-art results compared to existing reference-free evaluation scores. Our source code and trained models are publicly available at: {\url{https://github.com/aimagelab/bridge-score}}.
\end{abstract}

\section{Introduction\vspace{-0.1cm}} 
\label{sec:intro}

\begin{figure}[t]
    \centering
    \includegraphics[width=\linewidth]{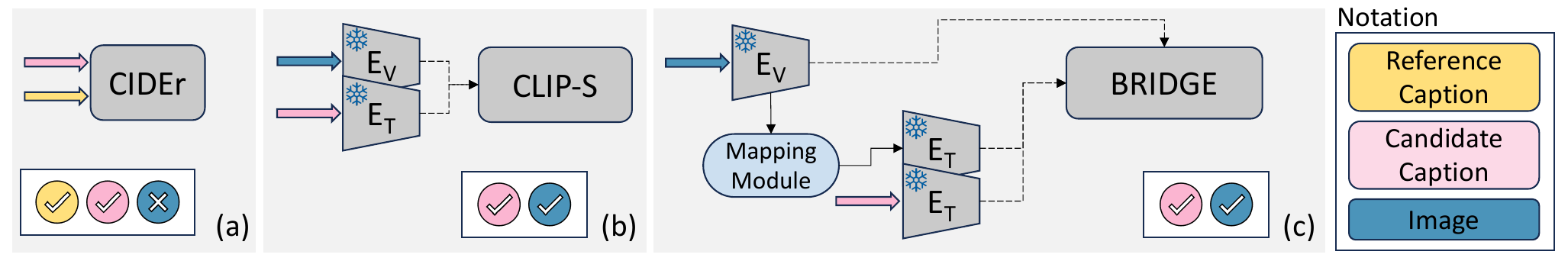}
    \vspace{-.5cm}
    \caption{Comparison between different captioning evaluation approaches: (a) CIDEr~\cite{vedantam2015cider} scores candidate and reference captions without considering the input image; (b) CLIP-Score~\cite{hessel2021clipscore} compares text and images using global vectors in a shared embedding space; (c) our \ours, internally builds multimodal pseudo-captions by translating fine-grained image features into pseudo-tokens thanks to a mapping module.}
    \label{fig:first}
    \vspace{-0.4cm}
\end{figure}

The objective of image captioning is to produce natural language descriptions conditioned on input images, that closely resemble human language and align to human intentions. As such, the captioning task involves the recognition and understanding of the visual content of the image, including fine-grained elements such as objects, attributes, and their relationships. 
Advances in training methodologies and architectures have contributed to the progress in the field, significantly improving the generation quality. Recent innovations include fully-attentive models~\cite{cornia2020meshed,herdade2019image,cornia2020smart}, improved connections between visual and textual modalities~\cite{cornia2020meshed,pan2020x}, and the incorporation of objects and tags at an architectural level~\cite{anderson2018bottom,li2020oscar,zhang2021vinvl}. Additionally, there has been a notable focus on increasing the robustness of cross-modal features~\cite{sarto2022retrieval,li2022comprehending,barraco2023little}, which consequently can increase description accuracy.

As constant improvements are made on the generation side, it becomes crucial to enhance the evaluation process as well.  
In this regard, image captioning evaluation aims to assess the quality of a generated caption given an image and, potentially, human-written reference captions. 
However, it is important to note that obtaining these reference captions can often be challenging and expensive, adding complexity to the evaluation process. 
Despite the recent advancements in captioning capabilities, standard automatic evaluation metrics have mainly relied on translation metrics~\cite{papineni2002bleu,lin2004rouge,banerjee2005meteor} or text-only ones~\cite{vedantam2015cider,spice2016,zhang2019bertscore} which often fall short in capturing aspects such as grammatical correctness, semantic relevance, and specificity. These limitations are worsened by the limited coverage of image content in available references, resulting in inaccurate penalties when generated captions accurately describe novel elements not mentioned in the references.

In response to these limitations, advanced metrics aligning visual and textual data have emerged~\cite{lee2020vilbertscore,lee2021umic,hessel2021clipscore,kim2022mutual}. Notably, recent metrics leverage the CLIP embedding space~\cite{radford2021learning}, which shows a strong correlation with human judgment. 
Despite the effectiveness of contrastive-based embedding spaces, metrics based on dual-encoder architectures tend to focus on global alignment between an image and its caption, often ignoring fine-grained details or penalizing hallucinations.

Following this insight, in this paper, we introduce \ours, a novel learnable and \textit{reference-free} image captioning metric that enhances the alignment of more fine-grained visual features. Specifically, our model provides a pre-trained dual-encoder architecture with a mapping module designed to effectively exploit visual cues (Fig.~\ref{fig:first}). This is done by internally creating multimodal pseudo-captions, containing both textual and dense visual features. 
The process for building these pseudo-captions involves the creation of a template caption, which focuses on the syntactical structure of the scene, and a mapping module. The latter refines the template caption by enriching it with more fine-grained visual features about the subjects depicted in the image. Subsequently, the overall model is trained with a combination of contrastive losses which promote multimodal alignment.

Experiments are conducted on a variety of datasets for image captioning annotated with human rankings, including Flickr8k-Expert, Flickr8k-CF~\cite{hodosh2013framing}, Composite~\cite{aditya2015images}, and Pascal-50S~\cite{vedantam2015cider}. Through comprehensive analysis, we demonstrate the efficacy of the proposed metric and show its ability to overcome the limitations of existing state-of-the-art reference-free alternatives. Additionally, we evaluate its sensitivity to object hallucination by conducting experiments on the FOIL dataset~\cite{shekhar2017foil}. Overall, \ours outperforms previous metrics and showcases superior performance compared to CLIP-Score~\cite{hessel2021clipscore} and PAC-Score~\cite{sarto2023positive}.

\tit{Contributions} In summary, our contributions are as follows: 
\begin{itemize}[noitemsep,topsep=0pt]
    \item We tackle the limitations of existing image captioning metrics by proposing the first learnable reference-free metric, termed as \ours, that focuses on more fine-grained visual features. Our proposal integrates a dual-encoder architecture with a mapping module which is in charge of producing multimodal pseudo-captions that combine text and richer dense visual features.
    \item Multimodal pseudo-captions are built by creating template captions that describe the scene from a syntactical point of view. These templates are subsequently enriched with fine-grained features through a mapping network.
    \item Experiments, carried out on different datasets with human preferences, demonstrate a higher degree of correlation with respect to existing metrics. We also evaluate the sensitivity of the proposal to objects hallucinations.
\end{itemize}
\section{Related Work\vspace{-0.1cm}}
\label{sec:related}

\tinytit{Classical Reference-based Captioning Metrics}
Several widely used captioning evaluation metrics were originally developed in the context of NLP tasks and rely on n-gram matching techniques. Among these classical metrics, BLEU~\cite{papineni2002bleu} is designed to focus on precision and incorporates a penalty for sentence brevity. METEOR~\cite{banerjee2005meteor}, instead, combines precision and recall to evaluate the quality of captions, while others, such as ROUGE~\cite{lin2004rouge}, were initially born for summarization tasks and later adapted to image captioning. More recently, two metrics tailored for visual captioning have emerged: CIDEr~\cite{vedantam2015cider}, which measures n-gram similarity and is based on TF-IDF, and SPICE~\cite{spice2016}, which quantifies graph-based similarity through scene graphs constructed from candidate and reference captions. Overall, focusing on textual-level comparisons, these metrics assume that human-written references accurately represent the image content. 

\tit{Learnable Captioning Metrics}
With the rise of large pre-trained models, image captioning evaluation now frequently exploits these models to compare textual-only~\cite{zhang2019bertscore,yi2020improving} or visual-textual~\cite{jiang2019tiger,jiang2019reo,lee2020vilbertscore,wang2021faier,lee2021umic,hessel2021clipscore} contents. Notably, the BERT score and its improved version use pre-trained BERT embeddings to compare word tokens in generated and ground-truth sentences.

Some metrics, like BLEU and CIDEr, rely solely on text matching between reference captions and machine-generated captions, potentially introducing bias in evaluations due to non-accurate reference captions. To mitigate these issues, alternative solutions leverage the multimodal nature of vision-and-language models. As an example, TIGEr~\cite{jiang2019tiger} considers the similarities between words and image regions and assesses how well machine-generated captions represent image content and their alignment with human-generated captions.

In contrast, other approaches~\cite{lee2020vilbertscore,hessel2021clipscore,kim2022mutual} leverage web-scale vision-and-language models such as VilBERT~\cite{lu2019vilbert} and CLIP~\cite{radford2021learning} for more robust metrics. 
For example, in~\cite{kim2022mutual},  CLIP visual-textual features are used to compute negative Gaussian cross-mutual information. Other works, instead, have employed diffusion models in text-only tasks~\cite{zhu2023imagine} or exploited the zero-shot language modeling capabilities of large language models~\cite{chan2023clair} to evaluate candidate captions.

\tit{Reference-free Captioning Metrics}
While all aforementioned metrics rely on a set of ground-truth captions to compute the final score, a few attempts have been made to introduce reference-free evaluations, only taking into account the correlation of the candidate caption with the image. In this regard, Lee~\etal~\cite{lee2021umic} proposed to fine-tune the UNITER model~\cite{chen2020uniter} via contrastive learning to let the model to discriminate between positive and synthetically-generated negative captions. On a different line, Hessel~\etal~\cite{hessel2021clipscore} introduced the CLIP-Score, which only relies on a modified cosine similarity between image and candidate caption representations coming from the CLIP model. The recently proposed PAC-Score~\cite{sarto2023positive}, instead, builds upon the usage of CLIP but incorporates a fine-tuning phase with positive augmentation, further enhancing the accuracy of evaluation. In this paper, we follow this research path and propose a novel learnable, and reference-free evaluation metric that can effectively incorporate fine-grained visual features for evaluating the correlation of image-text pairs.

\section{BRIDGE for Captioning Evaluation\vspace{-0.1cm}}
\begin{figure*}[t]
    \centering
    \includegraphics[width=\linewidth]{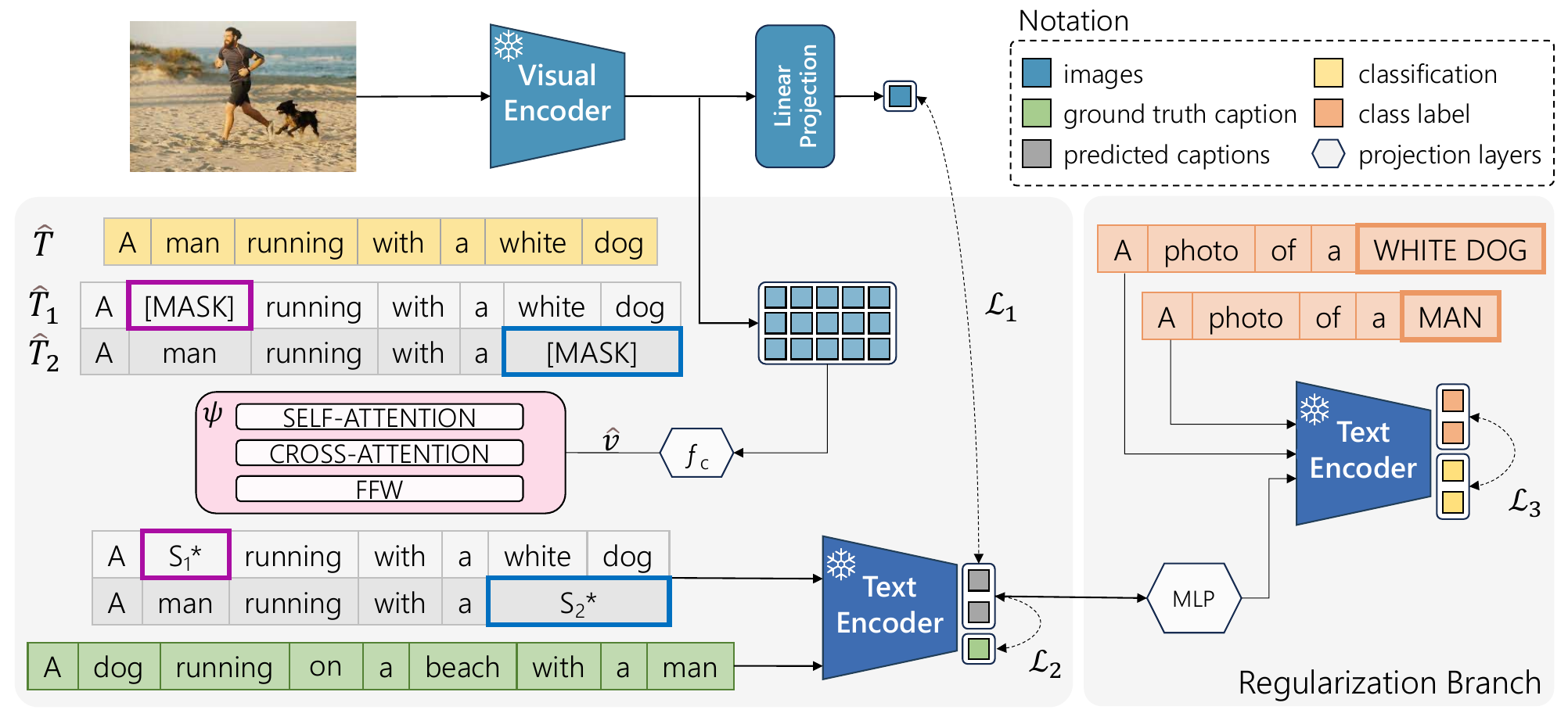}
    \vspace{-.5cm}
    \caption{Overview of the \ours evaluation approach. Starting from a template caption, a mapping network augments it with dense visual features, obtaining a pseudo-caption which is then used for computing image-text similarities.}
    \label{fig:model}
    \vspace{-0.35cm}
\end{figure*}

In the following, we present our reference-free captioning evaluation approach,
termed \textbf{\ours}. Our approach leverages a dual-encoder architecture, which comprises both a language encoder and a vision encoder. Given a frozen pre-trained model, we train a mapping module responsible for filling the holes of a masked template caption with \textit{pseudo language tokens} that are enriched with visual information. An overview of our model is depicted in Fig.~\ref{fig:model}.

\subsection{Preliminaries}
Our approach relies on CLIP (Contrastive Language–Image Pre-training)~\cite{radford2021learning}, a powerful vision and language model designed to align images and corresponding text captions within a shared embedding space. For a given input image $I$, the image encoder $E_V$ extracts the visual information $v = E_V(I) \in \mathbb{R}^d$. 
On the textual side, an input caption $T$ is tokenized and a textual representation is obtained by passing it through the textual encoder $E_T$, obtaining $t = E_T(T) \in \mathbb{R}^d$. Once the textual and visual features, $t$ and $v$ respectively, are projected in a common space, visual and textual inputs can be compared via cosine similarity. 

The relationships learned by CLIP can be exploited to build an image captioning evaluator. In CLIP-Score~\cite{hessel2021clipscore} the authors directly compare candidate captions and images in the embedding space and show that this achieves a good correlation with human judgments.
In detail, to assess the quality of a candidate generation, they feed both the image and the candidate caption through their respective feature extractors, and compute the cosine similarity of the resultant embeddings: 
\begin{equation}
\label{eq:clipscore}
   \text{CLIP-Score}(I, T) = w \cdot \max (\cos (v, t), 0),
\end{equation}
where $w$ is a rescaling factor employed to stretch the score distribution while ensuring the ranking results remain unchanged.

\subsection{Injecting Fine-Grained Visual Features} 
\label{sec:method}
Unlike CLIP-Score, our approach does not rely exclusively on global image descriptors for evaluating image-text alignments. Instead, we focus on employing stronger visual information. To do so, we draw inspiration from the Pic2Word approach~\cite{saito2023pic2word} and represent the input image through a multimodal pseudo-caption, an embedding representation that contains stronger visual elements.

\begin{figure}[t]
\centering
\includegraphics[width=\linewidth]{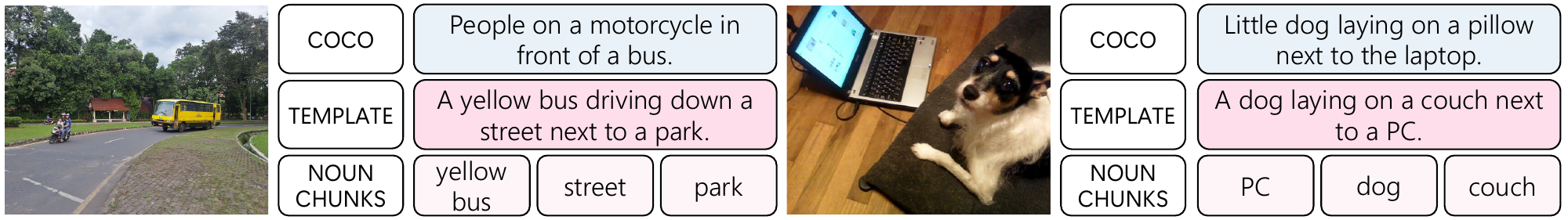}
\vspace{-.5cm}
\caption{COCO captions with template captions and associated noun chunks.} 
\label{fig:template_caps}
\vspace{-.4cm}
\end{figure}

\tit{Building Template Captions} In order to create multimodal pseudo-captions, we first build \textit{template captions} for a given input image. These are skeletal textual representations of the image, obtained by masking out all the relevant textual concepts from the descriptions generated by a captioner. Through these template captions, we aim to provide the model only with a templated textual structure which can then be filled with more fine-grained visual features. In particular, given an automatically generated caption describing the input image, such as \texttt{`A man running with a white dog'}, 
we remove the main subjects within the caption (\eg~\texttt{`man'} and \texttt{`white dog'}) and mask them with a \texttt{[MASK]} token. This will allow the model to fill in these gaps by incorporating more fine-grained features from the image encoder. Since a primary subject might be described by words other than just its corresponding nouns (\eg~adjectives), we utilize noun chunks. Fig.~\ref{fig:template_caps} reports template captions and corresponding noun chunks.

Given a sentence containing $N$ noun chunks, we independently encode them through the mapping network. To this aim, we replicate the template caption as many times as the number of noun chunks and mask a different noun chunk in each of the replicas. We thus obtain $N$ different versions of the template caption, each one masking a single noun chunk, for instance
\begin{gather}
   \nonumber
   \texttt{[`A [MASK] running with a white dog'},  \\
   \nonumber
\texttt{`A man running with a [MASK]'].}
\end{gather} 

\tit{Mapping with Fine-Grained Visual Features}
The above-described masked replicas are then fed to the mapping module $\psi$. Specifically, our approach exploits the visual information extracted from the visual encoder $E_V$ to enrich the replicas with the informative content of the image $I$. To get a more fine-grained representation of the image, we directly take the grid of features from the last layer, $\widehat{v}$. For instance, in the case of a ViT-B/32 backbone, this will have a shape of $50 \times d$, where $d$ is the dimensionality of the last embedding of the network. 

The mapping network is implemented as a stack of Transformer~\cite{vaswani2017attention} encoder layers interleaved with cross-attention layers. Its role is to refine each template captions with visual information. Since each template caption is processed independently, the mapping module returns a set of sequences, each with the same length as the corresponding input template caption.
From the output of the mapping module, we keep only the predictions for the masked tokens in each template caption and copy them back into the original templates.

Therefore, by providing a masked input template in the form $\hat{T_i} = \left[ w_1, ..., w_{j-1}, \texttt{MASK}, w_{j+1}, ..., w_T\right]$, where $\{w_j\}_j$ represent original tokens from the input caption, we obtain $\hat{T_i}^* = \left[ w_1, ..., w_{j-1}, \psi(\hat{T_i})_j, w_{j+1}, ..., w_T\right]$, where $\psi(\hat{T_i}, \widehat{v})_{j}$ represents the output of the mapping network at the position corresponding to the masked input token position. In the case of noun chunks consisting of more than one token, multiple consecutive tokens are replaced with the corresponding outputs from the mapping network.
By injecting the outputs of the mapping token into the initial template caption, we effectively complete the original templates with visually enriched vectors.
Notably, these newly generated pseudo-captions combine word sequences from the template captions with \textit{dense vectors} obtained by the mapping module. Consequently, they cannot be decoded as standard captions.
As a last step, the obtained pseudo-captions are fed into the pre-trained CLIP language encoder.

\subsection{Training Protocol} 
To train our mapping network, the loss is defined as a weighted version of the symmetric InfoNCE loss~\cite{oord2018representation}, where positive and negative items are weighted according to the number of noun chunks in each caption. 

Specifically, given a batch in the form $\mathcal{B}=\{ (I_i, T_i) \}_{i=1}^N$, where $I_i$ and $T_i$ represent image-caption pairs, each image $I_i$ is expanded in $N_i$ multimodal pseudo-captions as outlined above, where $N_i$ is the number of noun chunks in caption $T_i$. Further, let $\hat{t_{ij}}^*$ represent the embedding vector of the $j$-th pseudo-caption derived from the $i$-th image, $v_i$ the embedding vector of the $i$-th image and $t_i$ the embedding vector of the $i$-th ground-truth caption. Finally, let $M$ be the total number of noun chunks in the mini-batch, \ie~$M = \sum_{i=1}^N N_i$.

The first loss, denoted as $\mathcal{L}_1$, tries to align the pseudo-captions $\hat{t_{ij}}^*$ with the global visual features of the corresponding images $v_i$.
In addition to this loss term, we define a second loss component $\mathcal{L}_2$ that promotes the alignment between pseudo-captions $\hat{t_{ij}}^*$ and the textual feature vector of the ground-truth caption $t_i$ corresponding to the input image.
This ensure that pseduo-captions are aligned also on a textual space, in addition to being aligned in the image space. 
Additional details can be found in the supplementary materials.

\tit{Regularization Branch}
With the aforementioned loss terms, our objective is encouraging an association between each pseudo-caption and its corresponding image and caption. However, it is also important to differentiate each pseudo-caption from the others of the same image. To achieve this, we define a regularization loss which promotes a precise alignment between each pseudo-caption and the corresponding noun chunk.

First, we create prompts like ``\texttt{a photo of a $<$NOUNCHUNK$>$}'' and encode them with the text encoder $E_T$. In parallel, each pseudo-caption is fed into a dedicated multilayer perceptron (MLP) projection, which consist of two linear layers with a ReLU activation in between. 
Formally, the branch is defined as 
\begin{equation}
    \mathcal{C}(x) = \texttt{Linear}(\texttt{ReLU}(\texttt{Linear}(x))).
\end{equation} 

Since our goal is to emphasize each pseudo-caption's alignment with its corresponding noun chunk, we employ a regular contrastive loss $\mathcal{L}_3$
between the prompts mentioned earlier and the outputs of the projection branch.

Finally, the overall loss function we train \ours is defined as a weighted summation of two aforementioned losses, plus the regularization loss, as 
\begin{equation}
     \mathcal{L} = \lambda_1 \mathcal{L}_{1} + \lambda_2 \mathcal{L}_{2} + \lambda_3 \mathcal{L}_r.
\end{equation} 

\subsection{Inference and Score Computation} %
At inference time, given an image-candidate caption pair $(I, T)$, we extract all pseudo-captions from $I$ using our mapping network. Subsequently, we compute the mean pseudo-caption embedding as $t^* = \frac{1}{N} \hat{t_i}^*$, where $\hat{t_i}^*$ indicates the $i$-th pseudo-caption extracted from $I$ and $N$ here indicates the overall number of pseudo-captions associated with $I$.

At that point, given the visual embedding $v$ of the image and the embedding of the candidate caption $t$, the matching score between $I$ and $T$ is defined as
\begin{gather}
\label{eq:final_score}
    \text{\ours}(I, T) =  0.5 \cdot [ \text{CLIP-Score}(I, T)   + w \cdot
 \max (\cos(t^{*}, t ), 0)],
\end{gather}
where $\cos$ indicates the cosine similarity and $w$ is a constant scaling factor.

\section{Experimental Evaluation\vspace{-0.1cm}}
\label{sec:experiments}

\subsection{Implementation Details}
\tinytit{Architecture and Training Details} 
Building upon prior research~\cite{hessel2021clipscore,kim2022mutual,shi2022emscore}, we use either CLIP~\cite{radford2021learning} ViT-B/32 or ViT-L/14 as backbone for the visual and textual encoder. The mapping module is composed of two Transformer layers and is trained on the COCO dataset~\cite{lin2014microsoft}, which contains more than 120k images annotated with five captions. In particular, we employ the splits introduced by Karpathy~\etal~\cite{karpathy2015deep}, where 5,000 images are used for both validation and testing and the rest for training. To map the grid visual features to an embedding space of dimension 512, we employ a simple linear projection. For the regularization branch, we utilize a two-layer multi-layer perceptron. 

During training, we use AdamW~\cite{loshchilov2019decoupled} as optimizer with a learning rate equal to 0.0001 and a batch size of 256. 
The $\lambda_1$, $\lambda_2$, and $\lambda_3$ values are selected with a grid search, choosing the combination that provides the best validation loss. Specifically, we set both $\lambda_1$ and $\lambda_3$ to 0.01, while $\lambda_2$ is set to 1.0. The training stage lasts around one day on a single A100 GPU.

\tit{Template Caption Generation}
The template captions used as input for the mapping module are generated using the BLIP model~\cite{li2022blip}. In particular, we use the ViT-L/14 version pre-trained on 129M image-text pairs and finetuned on the COCO dataset. After this generation phase, the primary subjects of the template sentences are extracted by using the NLTK library~\cite{bird2009natural}.
During training, two noun chunks are randomly chosen from the set identified during the extraction step. In the evaluation phase, otherwise, all identified noun chunks are included.

\subsection{Datasets}
To evaluate the correlation of the proposed metric with human ratings, we conduct experiments on the Flickr8k-Expert, Flickr8k-CF, Composite, and Pascal50-S datasets~\cite{hodosh2013framing,aditya2015images,vedantam2015cider}. In addition, for detecting hallucinations in textual sentences, we extend our analysis to the FOIL dataset~\cite{shekhar2017foil}. Except for Pascal-50S and FOIL in where accuracy scores are used, evaluation on all other datasets relies on Kendall $\tau_b$, Kendall $\tau_c$, and Spearman $\rho$ correlation scores.

\tit{Flickr8k-Expert and Flickr8k-CF~\cite{hodosh2013framing}} These datasets consist of image-caption pairs with corresponding human ratings. Specifically, Flickr8k-Expert comprises 17k expert annotations for visual-textual pairs, with a total of 5,664 different images. Each pair receives a score ranging from 1 (lack of correlation) to 4 (accurate depiction), where 1 indicates a lack of correlation between the caption and the image, and 4 indicates an accurate depiction of the image without errors. On the other hand, Flickr8k-CF is composed of 145k binary quality judgments, collected from CrowdFlower, for 48k image-caption pairs containing 1,000 unique images. Each pair is annotated with at least three binary scores, where ``yes'' denotes that the caption correlates with the image. To measure the alignment with human judgment, we compute the mean proportion of ``yes'' annotations as the score for each pair.

\tit{Composite~\cite{aditya2015images}} It contains 12k human ratings for image-caption pairs including a combination of images taken from COCO~\cite{lin2014microsoft} (2,007 images), Flickr8k~\cite{hodosh2013framing} (997 images), and Flickr30k~\cite{young2014image} (991 images). In this dataset, human evaluators assess each image-caption pair, assigning a score within the range of 1 to 5 to estimate the correspondence of the caption with the associated image.

\tit{Pascal50-S~\cite{vedantam2015cider}} It presents pairwise preference judgments between two captions. Overall, the dataset consists of 4,000 sentence pairs, each of them associated with an image from the UIUC Pascal sentence dataset~\cite{rashtchian2010collecting}. Each pair is associated with 48 human judgments, where each evaluation indicates which sentence better describes the given image. The sentence pairs are categorized into four groups: (i) both human-written and correct captions (HC), (ii) both human-written captions where one is correct and the other is wrong (HI), (iii) both correct captions but one written by humans and the other machine-generated (HM), (iv) both machine-generated and correct captions (MM).

\tit{FOIL~\cite{shekhar2017foil}} The dataset comprises image-caption pairs from the COCO dataset~\cite{lin2014microsoft}. In this scenario, captions undergo perturbation by generating modified versions that closely resemble the originals but introduce a single error, referred to as ``foil word''. For a fair comparison, we select the subset of the validation set that does not overlap with the portion of COCO used during training, resulting in 8,000 images, each paired with a foil-correct textual counterpart.

\begin{table}[t]
    \centering
\caption{Ablation study results. US indicates the number of pseudo tokens.}
\vspace{-0.1cm}
\small
\centering
\setlength{\tabcolsep}{.2em}
\resizebox{\linewidth}{!}{
\begin{tabular}{lc ccc c ccc c c}
\toprule
& & \multicolumn{3}{c}{\textbf{Flickr8k-Expert}} & & \multicolumn{3}{c}{\textbf{Flickr8k-CF}} & & \textbf{Pascal-50S} \\
\cmidrule{3-5} \cmidrule{7-9} \cmidrule{11-11}
& & Kend. $\tau_b$ & Kend. $\tau_c$ & Spear. $\rho$ & & Kend. $\tau_b$ & Kend. $\tau_c$ & Spear. $\rho$ & & Accuracy \\
\midrule
\rowcolor{TitleColor} \multicolumn{3}{l}{Architectural Components} & & & & & & & & \\
\hspace{0.3cm}w/o mapping module (\ie~MLP) & & 53.1 & 53.5 & 65.3 & & 35.5 & 18.3 & 43.5 & & 81.9\\
\hspace{0.3cm}w/o template captions  & &  53.7 & 54.1 & 66.0 & &  35.5 & 18.4 & 43.6 & &  82.5 \\
\hspace{0.3cm}w/o regularization branch  & &  54.1 & 54.5 & 66.5 & &  35.7 & 18.5 & 43.6  & &  82.7 \\
\midrule
\rowcolor{TitleColor} \multicolumn{3}{l}{Score Formulation} & & & & & & & & \\
\hspace{0.3cm}w/o textual similarity  & & 51.1 & 51.2 & 63.0  & & 34.4 & 17.7 & 30.5  & & 80.9 \\
\hspace{0.3cm}w/o visual similarity & &  53.8 & 54.2 & 66.0  & & 35.4 & 18.3 & 43.7  & & 81.9 \\
\midrule
\rowcolor{TitleColor} \multicolumn{3}{l}{Pseudo-token Size} & & & & & & & & \\
\hspace{0.3cm}w/ $\text{US}=1$ & & 54.1 & 54.5 & 66.4 & &  35.1 & 17.1 & 30.3 & &  82.6 \\
\hspace{0.3cm}w/ $\text{US}=2$ & & 54.1 & 54.5 & 66.4 & &  \textbf{36.1} & \textbf{18.7} & 44.4 & &  \textbf{82.8} \\
\hspace{0.3cm}w/ $\text{US}=4$ & &  54.3 & 54.7 & 66.6 & &  35.4 & 18.3 & 43.8 & &  82.4 \\
\hspace{0.3cm}w/ $\text{US}=8$ & &  54.0 & 54.4 & 66.3 & &  35.9 & 18.6 & \textbf{44.5} & &  81.7 \\
\rowcolor{LightCyan}
\textbf{\ours} ($\text{US}=3$) & &  \textbf{54.4} & \textbf{54.8} & \textbf{67.7} & &  \textbf{36.1} & \textbf{18.7} & \textbf{44.5} & &  82.6 \\
\bottomrule
\end{tabular}
}
\vspace{-0.4cm}
\label{tab:ablation}
\end{table}

\subsection{Ablation Studies and Analysis}
To evaluate the effectiveness of our metric, we start by analyzing variations of our main architectural components. Then, we assess the impact of caption templates in our score formulation. All these experiments are performed using CLIP ViT-B/32 as backbone and reported in Table~\ref{tab:ablation}.

\tit{Contribution of Architectural Components}
We first investigate the performance of the most straightforward implementation of a mapping module, structured as a two-layer MLP following~\cite{saito2023pic2word}. We also validate the importance of the template captions through a model variant in which a set of learnable tokens \texttt{S$^*$} serves as input for the mapping module, without relying on template captions. In both variants, given the absence of template captions, we construct a template such as \texttt{`a photo of S$^*$'} and extract its features using the CLIP text encoder. In the Table, it can be seen that, regardless of any architectural changes, it is important to provide a simple sentence structure to the mapping module to achieve competitive performance.

In addition to these baselines, we devise a variant to analyze the contribution of the regularization branch. In this setting, we employ template captions as input for the mapping module, resulting in a substantial improvement of +1.0 Kendall $\tau_b$ and +0.8 accuracy points compared to the MLP variant, respectively on the Flickr8k-Expert and on the Pascal-50S dataset. When introducing the regularization branch (\ie~the complete BRIDGE architecture), further enhancements can be observed especially on the Flickr8k-CF with an improvement of +0.4 points in terms of the Kendall $\tau_b$ correlation score.

We also emphasize the importance of each component in our score formulation. Specifically, we present correlation results when employing only the visual similarity within our architecture, which is the original CLIP-Score formulation. As observed, performance drops drastically when relying only on visual information. A less significant drop is observed when employing only textual similarity.

As an additional analysis, we report the effect of changing the number of the pseudo tokens for each noun chunk, denoting it as unit size ($\text{US}$). Specifically, we compute the scores employing $\text{US}=1,2,3,4,8$. From the results, it can be seen that $\text{US}=3$ generally leads to the best performance across nearly all evaluation metrics. This configuration is used in all experiments reported in the paper.

\begin{table}[t]
\centering
\begin{minipage}{0.55\textwidth}
\caption{Impact of different template captions.}
\vspace{-0.1cm}
\small
\centering
\setlength{\tabcolsep}{.45em}
\resizebox{\linewidth}{!}{
\begin{tabular}{lc cc cc c}
\toprule
& & \textbf{Expert} & & \textbf{CF} & & \textbf{Pascal-50S} \\
\cmidrule{3-3} \cmidrule{5-5} \cmidrule{7-7}
 & & Kend. $\tau_b$ & & Kend. $\tau_b$ & & Acc.  \\
\midrule
\rowcolor{TitleColor} \multicolumn{3}{l}{Transformer Templates} & & & & \\
\hspace{0.3cm}w/o mapping module & &  46.1 & &  31.6 & & 80.4 \\
\rowcolor{LightCyan}
& &  \textbf{54.0} & &  \textbf{35.9} & & \textbf{82.7}  \\
\rowcolor{LightCyan}
\hspace{0.3cm}\multirow{-2}{*}{\textbf{\ours}} & & (\textcolor{magenta}{+7.9}) & &  (\textcolor{magenta}{+4.3}) & & (\textcolor{magenta}{+2.3}) \\
\midrule
\rowcolor{TitleColor} \multicolumn{3}{l}{BLIP Templates} & & & & \\
\hspace{0.3cm}w/o mapping module & & 48.6 & &  33.8 & &  81.0 \\
\rowcolor{LightCyan}
 & &  \textbf{54.4} & &  \textbf{36.1} & &  \textbf{82.6} \\
\rowcolor{LightCyan}
\hspace{0.3cm}\multirow{-2}{*}{\textbf{\ours}} & & (\textcolor{magenta}{+5.8}) & &  (\textcolor{magenta}{+2.3}) & & (\textcolor{magenta}{+1.6}) \\
\midrule
\rowcolor{TitleColor} \multicolumn{3}{l}{BLIP-2 Templates} & & & & \\
\hspace{0.3cm}w/o mapping module & &  49.4 & &  34.2 & &  82.4 \\
\rowcolor{LightCyan}
 & & \textbf{54.4} & &  \textbf{36.2} & &  \textbf{82.9} \\
\rowcolor{LightCyan}
\hspace{0.3cm}\multirow{-2}{*}{\textbf{\ours}} & & (\textcolor{magenta}{+5.0}) & &  (\textcolor{magenta}{+2.0}) & & (\textcolor{magenta}{+0.5}) \\
\bottomrule
\end{tabular}
}
\label{tab:templaten}
\end{minipage}
\hspace{0.08cm}
\begin{minipage}{0.42\textwidth}
\centering
\includegraphics[width=0.95\linewidth]{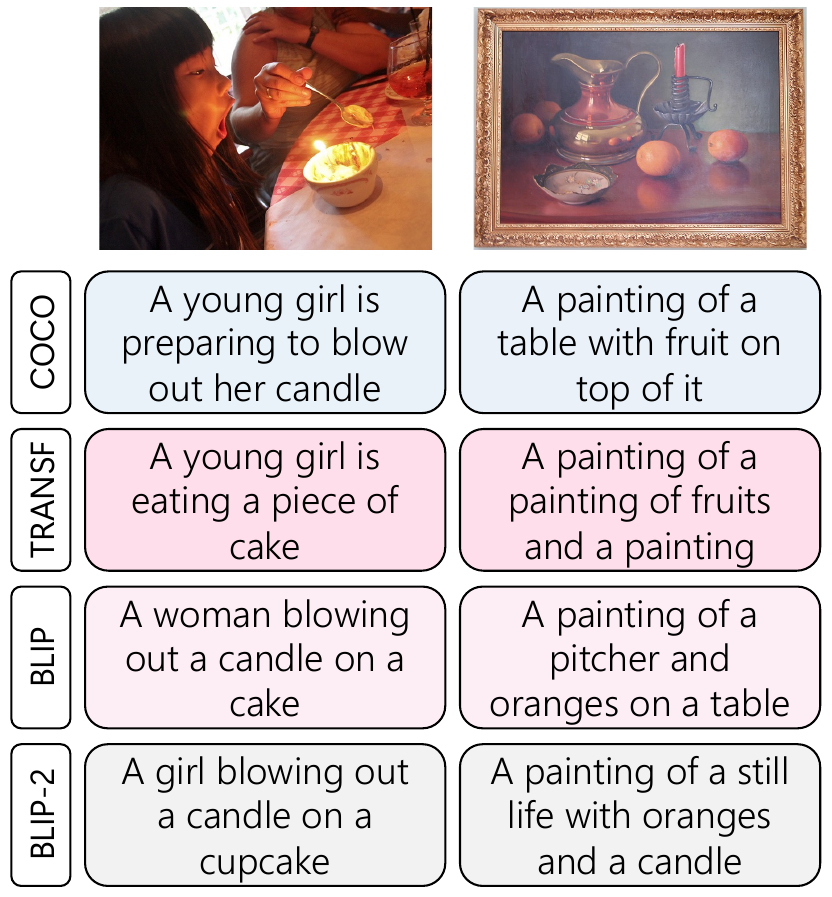}
\end{minipage}
\vspace{-0.4cm}
\end{table}

\tit{Analysis on Caption Templates} 
We analyze the effect of changing the initial template captions for our model. Specifically, we employ template captions generated by a conventional Transformer-based captioner that uses CLIP features as input and is trained only on the COCO dataset, as well as those generated by BLIP~\cite{li2022blip}, and BLIP-2~\cite{li2023blip2}. Notably, we select template captions of different quality based on both standard metric evaluations and correlations with human judgment. To assess the quality of the raw generated templates, we observe that the CIDEr score of these models on the COCO test set is equal to 114.2, 131.4, and 145.8 respectively for the standard Transformer model, BLIP, and BLIP-2. Note that all models were trained with cross-entropy loss only. Results on the Flickr8k-Expert, Flickr8k-CF, and Pascal-50S datasets are reported in Table~\ref{tab:templaten}. To qualitatively validate the generated templates, we include sample captions generated by the three models compared to a ground-truth caption from the COCO test set. Specifically, captions generated by BLIP-2 are generally more detailed and effectively describe the visual content of the input image compared to those generated by BLIP and, notably, the standard Transformer model.

For each caption template source, we compute the correlation scores when the mapping module is disabled and the captions are directly fed to the text encoder. We compare it with our standard \ours score, considering the different caption templates. Across all datasets, it is evident that directly using the caption templates as input to the text encoder leads to poor performance. 
This highlights the intended flexibility of template captions as skeletal representations, allowing the model to enhance them with fine-grained visual features.  

In fact, starting from these simple template captions and following our approach, we achieve improvements of +5.8 and +2.3 Kendall $\tau_b$ points and +1.6 accuracy points, respectively, on the Flickr8k-Expert, Flickr8k-CF, and Pascal50-S datasets when using BLIP caption templates.
The overall best results are with captions from the BLIP-2 model, confirming that better templates can indeed lead to improved results. However, even when using lower-quality captions, the final correlation results are very close to those obtained with higher-quality captions. This highlights the robustness of our metric to caption templates of varying quality and that it is not necessary to rely on captions generated by large-scale captioners to achieve strong correlation scores.

\begin{table}[t]
\caption{Correlation scores on Flickr8k-Expert, Flickr8k-CF, and Composite~\cite{hodosh2013framing,aditya2015images}.}
\vspace{-0.1cm}
\small
\centering
\setlength{\tabcolsep}{.18em}
\resizebox{\linewidth}{!}{
\begin{tabular}{lc ccc c ccc c ccc}
\toprule
& & \multicolumn{3}{c}{\textbf{Flickr8k-Expert}} & & \multicolumn{3}{c}{\textbf{Flickr8k-CF}} & & \multicolumn{3}{c}{\textbf{Composite}} \\
\cmidrule{3-5} \cmidrule{7-9} \cmidrule{11-13}
& & Kend. $\tau_b$ & Kend. $\tau_c$ & Spear. $\rho$ & & Kend. $\tau_b$ & Kend. $\tau_c$ & Spear. $\rho$ & & Kend. $\tau_b$ & Kend. $\tau_c$ & Spear. $\rho$ \\
\midrule
\rowcolor{TitleColor} \multicolumn{3}{l}{Reference-based metrics} & & & & & & & & & &\\
BLEU-4~\cite{papineni2002bleu} & & 30.6 & 30.8 & 38.7 & & 16.9 & 8.7 & 21.0 & & 28.3 & 30.6 & 38.1\\
METEOR~\cite{banerjee2005meteor} & & 41.5 & 41.8 & 51.9 & & 22.2 & 11.5 & 27.6 & & 36.0 & 38.9 & 48.1\\
CIDEr~\cite{vedantam2015cider} & & 43.6 & 43.9 & 54.3 & & 24.6 & 12.7 & 30.5 & & 34.9 & 37.7 & 47.0\\
SPICE~\cite{spice2016} & & 51.7 & 44.9 & 55.1 & & 24.4 & 12.0 & 31.3 & & 38.8 & 40.3 & 49.1\\
BERT-S~\cite{zhang2019bertscore} & & - & 39.2 & 50.3 & & 22.8 & - & - & & 39.9 & 30.1 & 48.6 \\
BERT-S++~\cite{yi2020improving} & & 48.1 & 46.7 & 56.9 & & - & - & - & & 42.3 & 44.9 & 52.1 \\
TIGEr~\cite{jiang2019tiger} & & 51.4 & 49.3 & 48.6 & & - & - & - & & 47.5 & 45.4 & 55.3 \\
ViLBERTScore~\cite{lee2020vilbertscore} & & 54.2 & 50.1 & 61.3 & & - & - & - & & 51.4 & 52.4 & 58.7 \\
MID~\cite{kim2022mutual} & & - & 54.9 & - & & 37.3 & - & - & & - & - & - \\
RefCLIP-S~\cite{hessel2021clipscore} & & 52.6 & 53.0 & 64.6 & & 36.4 & 18.8 & 44.7 & & 51.2 & 55.4 & 66.2 \\
RefPAC-S~\cite{sarto2023positive} & & {55.5} & {55.9} & 67.6 & & {37.6} & {19.5} & {46.2} & & {53.0} & {57.3} & 68.0 \\
\midrule
\midrule
\rowcolor{TitleColor} \multicolumn{3}{l}{Reference-free metrics} & & & & & & & & & &\\
UMIC~\cite{lee2021umic} & & - & 46.8 & - & & - & - & - & & - & - & - \\
\midrule
CLIP-S (ViT-B/32)~\cite{hessel2021clipscore} & & 51.1 & 51.2 & 63.0 & & 34.4 & 17.7 & 30.5 & & 49.8 & 53.8 & 64.3 \\
PAC-S (ViT-B/32)~\cite{sarto2023positive} & & 53.9 & 54.3 & 66.1 & & 36.0 & 18.6 & 44.4 & & \textbf{51.5} & \textbf{55.7} & \textbf{66.3} \\
\rowcolor{LightCyan}
\textbf{\ours} (ViT-B/32) & & \textbf{54.4} & \textbf{54.8} & \textbf{66.7} & & \textbf{36.1} & \textbf{18.7} & \textbf{44.5} & & {50.9} & {55.0} & {65.4} \\
\midrule
CLIP-S (ViT-L/14)~\cite{hessel2021clipscore} & & 52.6 & 53.0 & 64.7 & & {35.2} & 18.2 & {43.3} & & 51.3 & 55.4 & 65.9 \\
PAC-S (ViT-L/14)~\cite{sarto2023positive} & & 55.1 & 55.5 & 67.3 & & \textbf{36.8} & \textbf{19.0} & \textbf{45.3} & & {52.3} & {56.5} & {67.1} \\
\rowcolor{LightCyan}
\textbf{\ours} (ViT-L/14) & & \textbf{55.4} & \textbf{55.8} & \textbf{67.7} & & 36.3 & \textbf{19.0} & {44.7} & & \textbf{52.9} & \textbf{57.2} & \textbf{67.8} \\
\bottomrule
\end{tabular}
 }
\label{tab:flickr_composite}
\vspace{-0.4cm}
\end{table}

\subsection{Comparison with State-of-the-Art Captioning Metrics}

\tinytit{Evaluating Sample-Level Human Correlation}
We evaluate the sample-level human correlation on the Flickr8k~\cite{hodosh2013framing} and Composite~\cite{aditya2015images} datasets.
Following previous works~\cite{zhang2019bertscore,hessel2021clipscore}, we compute Kendall correlation scores in both $\tau_b$ and $\tau_c$ versions and also include the Spearman $\rho$ score. Results are reported in Table~\ref{tab:flickr_composite}, where we compare our proposed \ours metric against other reference-free evaluation scores like UMIC~\cite{lee2021umic}, CLIP-S~\cite{hessel2021clipscore}, and PAC-S~\cite{sarto2023positive}. Moreover, we also compare with standard captioning evaluation metrics (\ie~BLEU~\cite{papineni2002bleu}, METEOR~\cite{banerjee2005meteor}, CIDEr~\cite{vedantam2015cider}, and SPICE~\cite{spice2016}) and more recent reference-based solutions that exploit text-only or cross-modal learned embeddings, such as BERT-S~\cite{zhang2019bertscore}, BERT-S++~\cite{yi2020improving}, TIGEr~\cite{jiang2019tiger}, VilBERTScore~\cite{lee2020vilbertscore}, and MID~\cite{kim2022mutual}. For completeness, we also include the reference-based versions of CLIP-S and PAC-S, termed RefCLIP-S and RefPAC-S, which however are not directly comparable with our solution as both rely on a set of five reference captions.

As it can be seen, \ours outperforms other reference-free metrics in terms of correlation with human judgment, achieving the highest scores on almost all datasets. Specifically, compared to CLIP-S, \ours shows improvements in terms of Kendall $\tau_c$ of +3.6 and +2.8 points when using ViT-B/32 and ViT-L/14 backbone on the Flickr8k-Expert dataset. These improvements extend consistently across all datasets and correlation scores. Compared to PAC-S, \ours still achieves superior results on both Flickr8k-Expert and Flickr8k-CF, while performing on par on the Composite dataset (\ie~PAC-S achieves the best results when using ViT-B/32, while \ours outperforms PAC-S using ViT-L/14).

\tit{Discriminating Correct Captions}
We also evaluate the effectiveness of our metric on the PASCAL-50S dataset~\cite{vedantam2015cider}. 
In this context, instead of 
\begin{wraptable}{r}{0.58\textwidth}
\vspace{-0.9cm}
\caption{Accuracy results on Pascal-50S~\cite{vedantam2015cider} averaged over five random draws of reference captions. The $\dagger$ marker indicates scores from previous works.
}
\vspace{0.1cm}
\small
\centering
\setlength{\tabcolsep}{.3em}
\resizebox{0.93\linewidth}{!}{
\begin{tabular}{lc cccc c c}
\toprule
 & & HC & HI & HM & MM & & Mean \\
\midrule
\rowcolor{TitleColor} \multicolumn{8}{l}{Reference-based metrics}  \\
BLEU-4~\cite{papineni2002bleu} & & 60.3 & 93.1 & 85.7 & 57.0 & & 74.0 \\
METEOR~\cite{banerjee2005meteor} & & 66.0 & 97.7 & 94.0 & 66.6 & & 81.1 \\
CIDEr~\cite{vedantam2015cider} & & 66.5 & 97.9 & 90.7 & 65.2 & & 80.1 \\
BERT-S++$^\dagger$~\cite{yi2020improving} & & 65.4 & 98.1 & 96.4 & 60.3 & & 80.1 \\
TIGEr$^\dagger$~\cite{jiang2019tiger} & & 56.0 & 99.8 & 92.8 & 74.2 & & 80.7 \\
MID$^\dagger$~\cite{kim2022mutual} & & 67.0 & 99.7 & 97.4 & 76.8 & & 85.2 \\
RefCLIP-S~\cite{hessel2021clipscore} & & 64.9 & 99.5 & 95.5 & 73.3 & & 83.3 \\
RefPAC-S~\cite{sarto2022retrieval} & & 67.7 & 99.6 & 96.0 & 75.6 & & 84.7 \\
\midrule
\midrule
\rowcolor{TitleColor} \multicolumn{8}{l}{Reference-free metrics}  \\
CLIP-S (ViT-B/32)~\cite{hessel2021clipscore} & & 55.9 & 99.3 & 96.5 & 72.0 & & 80.9 \\
PAC-S (ViT-B/32)~\cite{sarto2023positive} & & \textbf{60.6} & 99.3 & 96.9 & 72.9 & & 82.4 \\
\rowcolor{LightCyan}
\textbf{\ours} (ViT-B/32) & & 59.4 & \textbf{99.4} & \textbf{97.5} & \textbf{74.0} & & \textbf{82.6} \\
\midrule
CLIP-S (ViT-L/14)~\cite{hessel2021clipscore} & & 57.0 & \textbf{99.6} & \textbf{96.7} & {73.5} & & 81.7 \\
PAC-S (ViT-L/14)~\cite{sarto2023positive} & & 59.5 & 99.4 & 95.8 & \textbf{74.7} & & 82.2 \\
\rowcolor{LightCyan}
\textbf{\ours} (ViT-L/14) & & \textbf{61.2} & \textbf{99.6} & {96.6} & 74.1 & & \textbf{82.9} \\
\bottomrule
\end{tabular}
}
\vspace{-0.8cm}
\label{tab:pascal}
\end{wraptable}
calculating correlation scores, we compute accuracy by determining, for each pair, the caption favored by the majority of human ratings as correct (with ties being resolved randomly). We then measure how frequently the evaluation metric assigns a higher score to the chosen caption. Following previous works~\cite{hessel2021clipscore}, we randomly select five reference captions from the set of 48 provided by the dataset and average the results over five distinct draws. Accuracy values are reported in Table~\ref{tab:pascal}. The results show that when using both ViT-B/32 and ViT-L/14, \ours consistently outperform CLIP-S across all categories, showcasing an average accuracy increase of +1.7 and +1.2 points, respectively. When comparing with PAC-S, our solution can better discriminate the correct captions on both backbones with an average accuracy increase of +0.2 and +0.7 using ViT-B/32 and ViT-L/14 respectively.

\begin{figure}[t]
  \begin{minipage}[b]{0.60\textwidth}
    \centering
    \includegraphics[width=\textwidth]{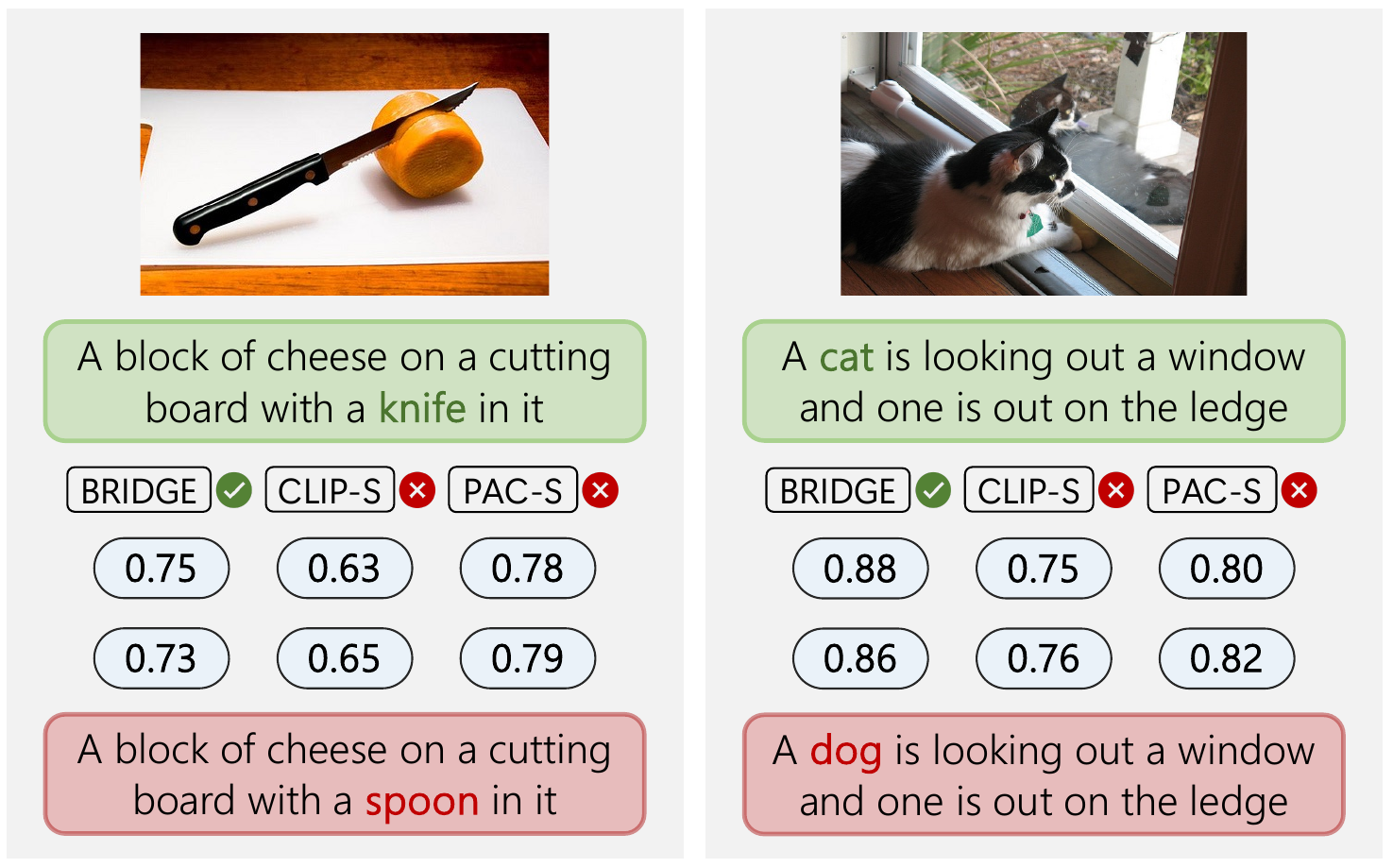}
    \vspace{-0.4cm}
    \captionof{figure}{Sample images from the FOIL dataset~\cite{shekhar2017foil} and corresponding scores generated by our proposed metric compared with CLIP-S and PAC-S.}
    \label{img:foil}
  \end{minipage}
  \hspace{0.01cm}
\begin{minipage}[b]{0.34\textwidth}
    \centering
    \captionof{table}{Accuracy results on the FOIL~\cite{shekhar2017foil} dataset.}
    \vspace{0.15cm}
    \setlength{\tabcolsep}{.4em}
    \resizebox{0.93\linewidth}{!}{
    \begin{tabular}[b]{lc cc c}
    \toprule
     & & \textbf{Acc.} \\
    \midrule
    \rowcolor{TitleColor} \multicolumn{3}{l}{Reference-based metrics}  \\
    BLEU-4~\cite{papineni2002bleu} & & 66.2 \\ 
    METEOR~\cite{banerjee2005meteor} & & 70.1 \\ 
    CIDEr~\cite{vedantam2015cider} & & 85.7 \\ 
    MID~\cite{kim2022mutual} & & 90.5 \\ 
    RefCLIP-S~\cite{hessel2021clipscore} & & 91.0 \\ 
    RefPAC-S~\cite{hessel2021clipscore} & & 93.7 \\ 
    \rowcolor{LightCyan}
    \midrule
    \midrule
    \rowcolor{TitleColor} \multicolumn{3}{l}{Reference-free metrics}  \\
    CLIP-S (ViT-B/32)~\cite{hessel2021clipscore} & & 87.2 \\ 
    PAC-S (ViT-B/32)~\cite{sarto2023positive} & & 89.9 \\
    \rowcolor{LightCyan}
    \textbf{\ours} (ViT-B/32) & & \textbf{91.5} \\ 
    \midrule
    CLIP-S (ViT-L/14)~\cite{hessel2021clipscore} & & 90.9 \\
    PAC-S (ViT-L/14)~\cite{sarto2023positive} & & 91.9 \\ 
    \rowcolor{LightCyan}
    \textbf{\ours} (ViT-L/14) & & \textbf{93.0} \\ 
    \bottomrule
    \end{tabular}
    \label{tab:foil}}
    \end{minipage}
    \vspace{-0.4cm}
\end{figure}

\tit{Object Hallucination Analysis} 
We then extend our analysis to the FOIL dataset~\cite{shekhar2017foil} for correctly identifying captions that may contain object hallucinations. 
Table~\ref{tab:foil} shows the accuracy results. As it can be seen, \ours outperforms CLIP-S and PAC-S, exhibiting an increase respectively of +4.3 and +1.6 points when using ViT-B/32 as backbone. Similar improvements can also be observed when employing more robust visual features. 
These results demonstrate the capabilities of our metric also to identify hallucinated objects correctly. In Fig~\ref{img:foil}, we present sample results comparing our metric with CLIP-S and PAC-S.

\subsection{System-level Correlation}
Finally, we delve into the efficacy of our proposed metric when evaluating popular existing captioning models. To this aim, we generate predictions of several state-of-the-art captioning models on the COCO test set, including Show and Tell and Show, Attend and Tell which are among the first image captioning models based on deep learning, Up-Down~\cite{anderson2018bottom}, SGAE~\cite{yang2019auto}, AoANet~\cite{huang2019attention}, $\mathcal{M}^2$ Transformer~\cite{cornia2020meshed}, X-Transformer~\cite{pan2020x} which all include region-based image features with either LSTM-based or Transformer-based language models, and the recently proposed COS-Net model~\cite{li2022comprehending} that incorporates CLIP features.
In addition to reporting evaluations on traditional captioning models, we also include recent LLM-based captioning models, including ZeroCap~\cite{tewel2022zerocap} and SmallCap~\cite{ramos2023smallcap}, which are based on GPT-2~\cite{radford2019language}, and MiniGPT-v2~\cite{chen2023minigpt}, BLIP-2~\cite{li2023blip2}, IDEFICS~\cite{laurenccon2023obelisc}, LLaVA-1.5~\cite{liu2024visual,liu2023improved}, and InstructBLIP~\cite{dai2023instructblip} which instead are based on larger-scale LLMs like Flan-T5~\cite{chung2022scaling}, Vicuna~\cite{vicuna2023}, or LLaMA~\cite{touvron2023llama,touvron2023llama2}.

The results are presented in Table~\ref{tab:captioners}, where we evaluate our \ours scores against both standard metrics, such as BLEU-4, METEOR, and CIDEr, and more recent ones like CLIP-S and PAC-S. Captioning models are compared to a human baseline, in which, for each sample, one human-annotated sentence (selected randomly from the five provided by the COCO dataset) serves as a candidate caption. 
As shown in the table, \ours can effectively evaluate human-annotated sentences which obtain a score similar to recent state-of-the-art captioning models such as COS-Net. This capability lacks in standard metrics such as METEOR and CIDEr which rank human captions lower than those generated by less-performing captioning models like Show, Attend, and Tell or Up-Down.

When considering the evaluation of captions generated by existing models, our metric shows a strong correlation with standard evaluation metrics when evaluating traditional image captioners or large-scale models that are fine-tuned on the COCO dataset like BLIP-2. When instead considering more recent approaches that are based on large language models and are not fine-tuned on COCO, \ours still recognizes the goodness of generated captions, raking InstructBLIP as the best-performing approach. This demonstrates the capabilities of our metric to correctly evaluate longer and more detailed captions which are typically generated by LLM-based multimodal models~\cite{caffagni2024r,li2024if,dong2024benchmarking} and that might be very different from captions contained in the COCO dataset.

\begin{table}[t]
\caption{Evaluation scores of traditional and LLM-based captioners on COCO test set ($\blacklozenge$: not trained/fine-tuned on COCO).}
\vspace{-0.2cm}
\small
\centering
\setlength{\tabcolsep}{.4em}
\resizebox{0.9\linewidth}{!}{
\begin{tabular}{cc lc ccc cccc}
\toprule
& & & & BLEU-4 & METEOR & CIDEr & CLIP-S & PAC-S & \textbf{\ours} \\
\midrule
\multirow{8}{*}{{\rotatebox[origin=c]{90}{\textit{Traditional}}}} & & Show and Tell~\cite{vinyals2015show} & & 31.4 & 25.0 & 97.2 & 0.572 &  0.772 & \cellcolor[rgb]{1,0.89,0.918} 0.788  \\
& & Show, Attend and Tell~\cite{xu2015show} & & 33.4 & 26.2 & 104.6 & 0.582 & 0.785 & \cellcolor[rgb]{1,0.89,0.918} 0.804 \\
& &  Up-Down~\cite{anderson2018bottom} & & 36.7 & 27.9 & 122.7 & 0.723 & 0.803 & \cellcolor[rgb]{1,0.89,0.918} 0.821 \\
& &  SGAE~\cite{yang2019auto} & & 39.0 & 28.4 & 129.1 & 0.734 & 0.812 & \cellcolor[rgb]{1,0.89,0.918} 0.833 \\
& &  AoANet~\cite{huang2019attention} & & 38.9 & 29.2 & 129.8 & 0.737 & 0.815 & \cellcolor[rgb]{1,0.89,0.918} 0.836 \\
& &  $\mathcal{M}^2$ Transformer~\cite{cornia2020meshed} & & 39.1 & 29.2 & 131.2 & 0.734 & 0.813 & \cellcolor[rgb]{1,0.89,0.918} 0.841  \\
& & X-Transformer~\cite{pan2020x} & & 39.7 & 29.5 & 132.8 & 0.610 & 0.812 & \cellcolor[rgb]{1,0.89,0.918} 0.845 \\
& &  COS-Net~\cite{li2022comprehending} & & 42.0 & 30.6 & 141.1 & 0.758 & 0.832 & \cellcolor[rgb]{1,0.89,0.918} 0.859 \\
\midrule
\multirow{7}{*}{{\rotatebox[origin=c]{90}{\textit{LLM-based}}}} & & ZeroCap$^\blacklozenge$~\cite{tewel2022zerocap} & & 2.3 & 10.1 & 15.1 & 0.810 & 0.816 & \cellcolor[rgb]{1,0.89,0.918} 0.862  \\
& & SmallCap~\cite{ramos2023smallcap} & & 37.0 & 27.9 & 119.7 & 0.748 & 0.826 & \cellcolor[rgb]{1,0.89,0.918} 0.847   \\
& & MiniGPT-v2~\cite{chen2023minigpt} & & 18.8 & 24.6 & 80.4 & 0.752 & 0.818 & \cellcolor[rgb]{1,0.89,0.918} 0.845  \\
& & BLIP-2~\cite{li2023blip2} & & \textbf{43.7} & \textbf{32.0} & \textbf{145.8} & 0.767 & \textbf{0.837} & \cellcolor[rgb]{1,0.89,0.918} 0.868\\
& & IDEFICS-9B$^\blacklozenge$~\cite{laurenccon2023obelisc} & & 4.3 & 19.1 & 50.0 & 0.740 & 0.786 & \cellcolor[rgb]{1,0.89,0.918} 0.838  \\
& & LLaVA-1.5-7B$^\blacklozenge$~\cite{liu2023improved} & & 8.1 & 28.0 & 69.6 & 0.784 & 0.809 & \cellcolor[rgb]{1,0.89,0.918} 0.867  \\
& & InstructBLIP-Flan-T5-XL$^\blacklozenge$~\cite{dai2023instructblip} & & 6.1 & 28.1 & 38.1 & \textbf{0.817} & \textbf{0.837} & \cellcolor[rgb]{1,0.89,0.918} \textbf{0.902} \\
\midrule 
& & \textit{Humans} & & - & \textit{24.1} & \textit{87.6} & {\textit{0.774}} &  \textit{0.823} & \cellcolor[rgb]{1,0.89,0.918} \textit{0.856} \\
\bottomrule
\end{tabular}
}
\label{tab:captioners}
\vspace{-0.35cm}
\end{table}

\section{Conclusion\vspace{-0.1cm}}
\label{sec:conclusion}
In this paper, we have presented a novel learnable, and reference-free image captioning metric that combines text and dense visual features. Our proposal, \ours, employs templated captions that are enriched with fine-grained visual cues thanks to a mapping network. Through experimental evaluation, we demonstrate that \ours outperforms existing reference-free metrics in terms of correlation with human judgment and sensitivity to hallucinated objects.

\section*{Acknowledgments}
We acknowledge the CINECA award under the ISCRA initiative, for the availability of high-performance computing resources. This work has been conducted under a research grant co-funded by Leonardo S.p.A. and supported by the PNRR-M4C2 (PE00000013) project ``FAIR - Future Artificial Intelligence Research'' and by the PRIN project ``MUSMA'' (CUP G53D23002930006 - M4C2 I1.1), both funded by EU - Next-Generation EU.

%
%
\bibliographystyle{splncs04}
\bibliography{main}

\newpage
\appendix
\section*{Supplementary Material}
\appendix

In the following, we present further experiments and analyses about the proposed \ours metric. Specifically, we provide a detailed description of the weighted contrastive loss function used to train our approach. Additionally, we report several supplementary qualitative results to support our findings.

\vspace{-0.1cm}

\section{Weighted Contrastive Loss\vspace{-0.1cm}}

In Section~\ref{sec:method} of the main paper, we state that we employ a weighted variant of the symmetric InfoNCE loss~\cite{oord2018representation}. 
Specifically, our method involves building mini-batches of multimodal pseudo-captions derived from a set of image-caption pairs. To recall the notation of that section, the mini-batched are in the form $\mathcal{B}=\{ (I_i, T_i) \}_{i=1}^N$, where $I_i$ and $T_i$ represent image-caption pairs. Each image $I_i$ is expanded in $N_i$ multimodal pseudo-captions, with $N_i$  representing the number of noun chunks in caption $T_i$. 
As in the main paper, we denote $\hat{t_{ij}}^*$ as the embedding vector of the $j$-th pseudo-caption derived from the $i$-th image, $v_i$ as the embedding vector of the $i$-th image, and $t_i$ as the embedding vector of the $i$-th ground-truth caption.
Finally, let $M$ be the total number of noun chunks in the mini-batch, \ie~$M = \sum_{i=1}^N N_i$.

The first weighted contrastive loss aligns the embeddings of the multimodal pseudo-captions with the global visual features of the corresponding images. This step ensures that each pseudo-caption is appropriately contextualized within the overall visual context. This loss is defined as a weighted version of the symmetric InfoNCE loss because positive and negative items are weighted according to the number of noun chunks in each caption. The rationale behind this choice is that captions having more noun chunks tend to have more visual variance. We therefore assign them a higher weight to promote the transfer of proper visual features. Formally, the loss is defined as follows
\begin{gather}
    \mathcal{L}_1 = -\frac{1}{M} \sum_{i=1}^N \sum_{j=1}^{N_i} \log \frac{\exp(\cos(v_i, \hat{t_{ij}}^*) / \tau)}{\sum_{k \neq i} N_k \cdot \exp(\cos(v_k, \hat{t_{ij}}^*) / \tau)} + \nonumber \\ 
    -\frac{1}{M} \sum_{i=1}^N \sum_{j=1}^{N_i} N_i \cdot \log \frac{\exp(\cos(v_i, \hat{t_{ij}}^*) / \tau)}{\sum_{k \neq i} \sum_{h=1}^{N_k} \exp(\cos(v_i, \hat{t_{hk}}^*) / \tau)}.
    \label{eq:l_1}
\end{gather}
Noticeably, differently from the standard InfoNCE loss, we also remove the positive item from the denominator of each loss component.

\begin{table}[t]
\small
\centering
\caption{Human correlation and accuracy scores changing the underlying backbone.}
\setlength{\tabcolsep}{.35em}
\resizebox{0.68\linewidth}{!}{
\begin{tabular}{lc cc cc cc cc}
\toprule
& & \textbf{Expert} & & \textbf{CF} & & \textbf{Pascal-50S} & & \textbf{FOIL}\\
\cmidrule{3-3} \cmidrule{5-5} \cmidrule{7-7} \cmidrule{9-9}
& & Kendall $\tau_b$ & & Kendall $\tau_b$ & & Accuracy & & Accuracy \\
\midrule
\rowcolor{TitleColor} \multicolumn{9}{l}{CLIP-based backbone} \\
\hspace{0.4cm}CLIP-S~\cite{hessel2021clipscore}  & &  51.1 & &  34.4 & &  80.9  & &  87.2 \\
\rowcolor{LightCyan}
\hspace{0.4cm}\textbf{\ours}  & &  \textbf{54.4} & &  \textbf{36.1} & &  \textbf{82.6}  & &  \textbf{91.5}\\
\midrule
\midrule
\rowcolor{TitleColor} \multicolumn{9}{l}{PAC-based backbone} \\
\hspace{0.4cm}PAC-S~\cite{sarto2023positive}  & &  53.9 & &  36.0 & &  82.4  & &  89.9\\
\rowcolor{LightCyan}
\hspace{0.4cm}\textbf{\ours}  & &  \textbf{54.8} & & \textbf{36.4} & & \textbf{82.7}  & &  \textbf{91.4} \\
\bottomrule
\end{tabular}
\vspace{-0.3cm}
}

\label{tab:diff_backbone}
\end{table}

In addition to the above-defined loss, we define a second loss component that promotes the alignment between pseudo-captions and the textual feature vector of the ground-truth caption corresponding to the input image. This makes sure that pseudo-captions are aligned also on a textual space, in addition to being aligned in the image space. Symmetrically to Eq.~\ref{eq:l_1}, this loss is defined as
\begin{gather}
    \mathcal{L}_2 = -\frac{1}{M} \sum_{i=1}^N \sum_{j=1}^{N_i} \log \frac{\exp(\cos(t_i, \hat{t_{ij}}^*) / \tau)}{\sum_{k \neq i} N_k \cdot \exp(\cos(t_k, \hat{t_{ij}}^*) / \tau)} + \nonumber \\ 
    -\frac{1}{M} \sum_{i=1}^N \sum_{j=1}^{N_i} N_i \cdot \log \frac{\exp(\cos(t_i, \hat{t_{ij}}^*) / \tau)}{\sum_{k \neq i} \sum_{h=1}^{N_k} \exp(\cos(t_i, \hat{t_{hk}}^*) / \tau)},
    \label{eq:l_2}
\end{gather}
where $t_i$ is the global textual feature vector of the ground-truth caption for $V_i$.

\section{Additional Experimental Results\vspace{-0.1cm}}

\tinytit{Effect of Changing the Underling Backbone}
The proposed \ours score is based on the standard CLIP model without fine-tuning its original weights. However, recent solutions like PAC-S~\cite{sarto2023positive} improve the performance of CLIP-S by fine-tuning the final projections of visual and textual encoders with curated data. In Table~\ref{tab:diff_backbone}, we assess whether applying the proposed \ours approach to the fine-tuned backbone employed in PAC-S can further improve its final results. Interestingly, \ours can not only enhance the results of a standard CLIP-based model but can also achieve improved correlation with human judgment when using the fine-tuned CLIP model presented in~\cite{sarto2023positive}, termed as PAC in the table. This further demonstrates the effectiveness of our evaluation score and its generalization capabilities when employing different backbones. 

\tit{Additional Ablation Studies}
In the upper part of Table~\ref{tab:ablation_supp}, we present the results across different datasets when employing the standard contrastive loss instead of considering the number of noun chunks in each caption. 
The results indicate that employing the standard loss does not enhance the final performance, thereby confirming the advantages of prioritizing captions that contain a greater number of noun chunks.

\begin{table}[t]
\small
\caption{Additional ablation study results on architectural design.}
\vspace{-0.2cm}
\centering
\setlength{\tabcolsep}{.25em}
\resizebox{0.7\linewidth}{!}{
\begin{tabular}{lc cc cc cc}
\toprule
& & \textbf{Expert} & & \textbf{CF} & & \textbf{Pascal-50S} \\
\cmidrule{3-3} \cmidrule{5-5} \cmidrule{7-7}
& & Kendall $\tau_b$ & & Kendall $\tau_b$ & & Accuracy  \\
\midrule
\rowcolor{TitleColor} \multicolumn{7}{l}{Loss Design}  \\
w/ standard contrastive losses & &  54.2 & & 35.6 & & 82.5\\
\rowcolor{TitleColor} \multicolumn{7}{l}{Architectural Design}  \\
w/ entire mapping module output & &  52.0 & &  34.5 & &  82.5 \\

w/ global features & &  53.9 & &  35.1 & &  82.3 \\
\midrule
\rowcolor{LightCyan}
\textbf{\ours} & &  \textbf{54.4} & &  \textbf{36.1} & &  \textbf{82.6} \\
\bottomrule
\end{tabular}
}
\label{tab:ablation_supp}
\vspace{-0.2cm}
\end{table}

\begin{figure}[t]
  \centering
 \resizebox{0.95\linewidth}{!}{
    \setlength{\tabcolsep}{.7em}
    \begin{tabular}{cc}
    \includegraphics[width=0.48\linewidth]{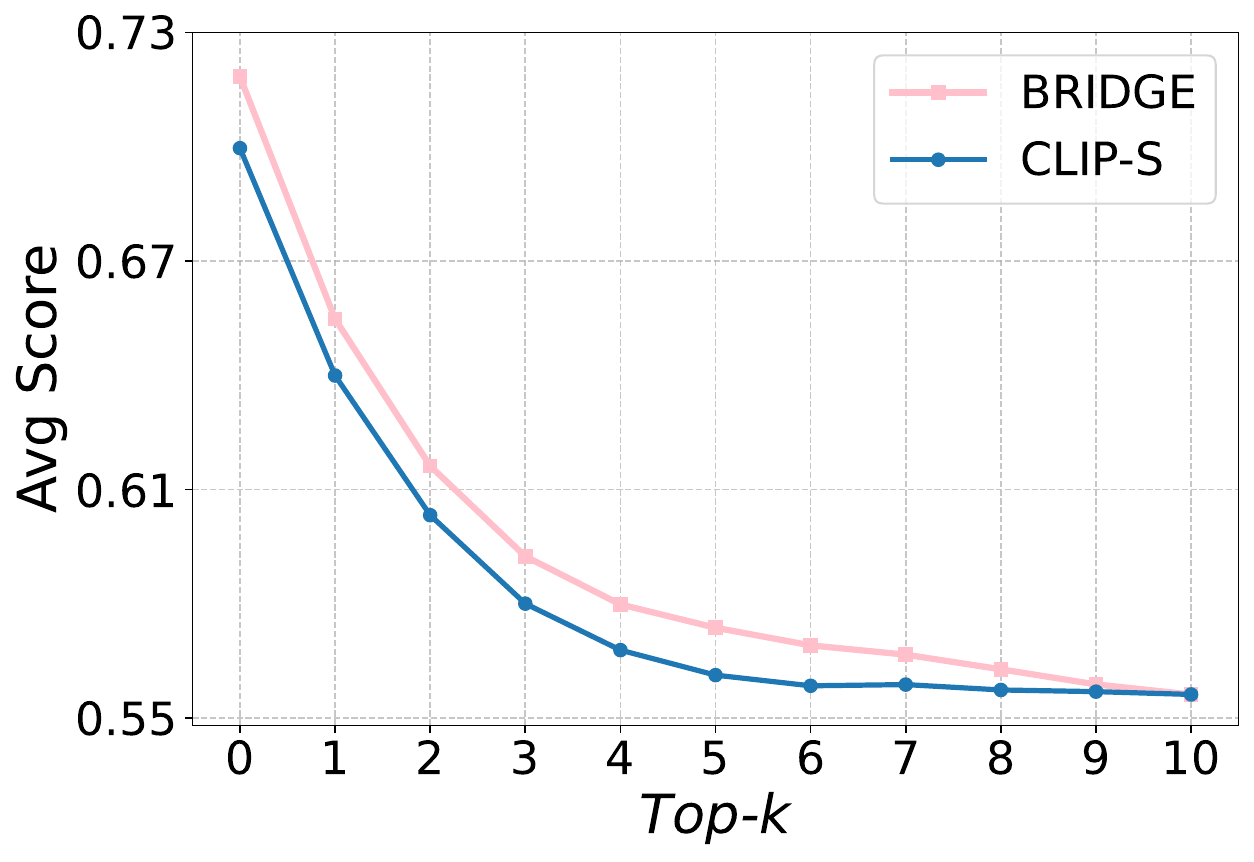} & 
    \includegraphics[width=0.48\linewidth]{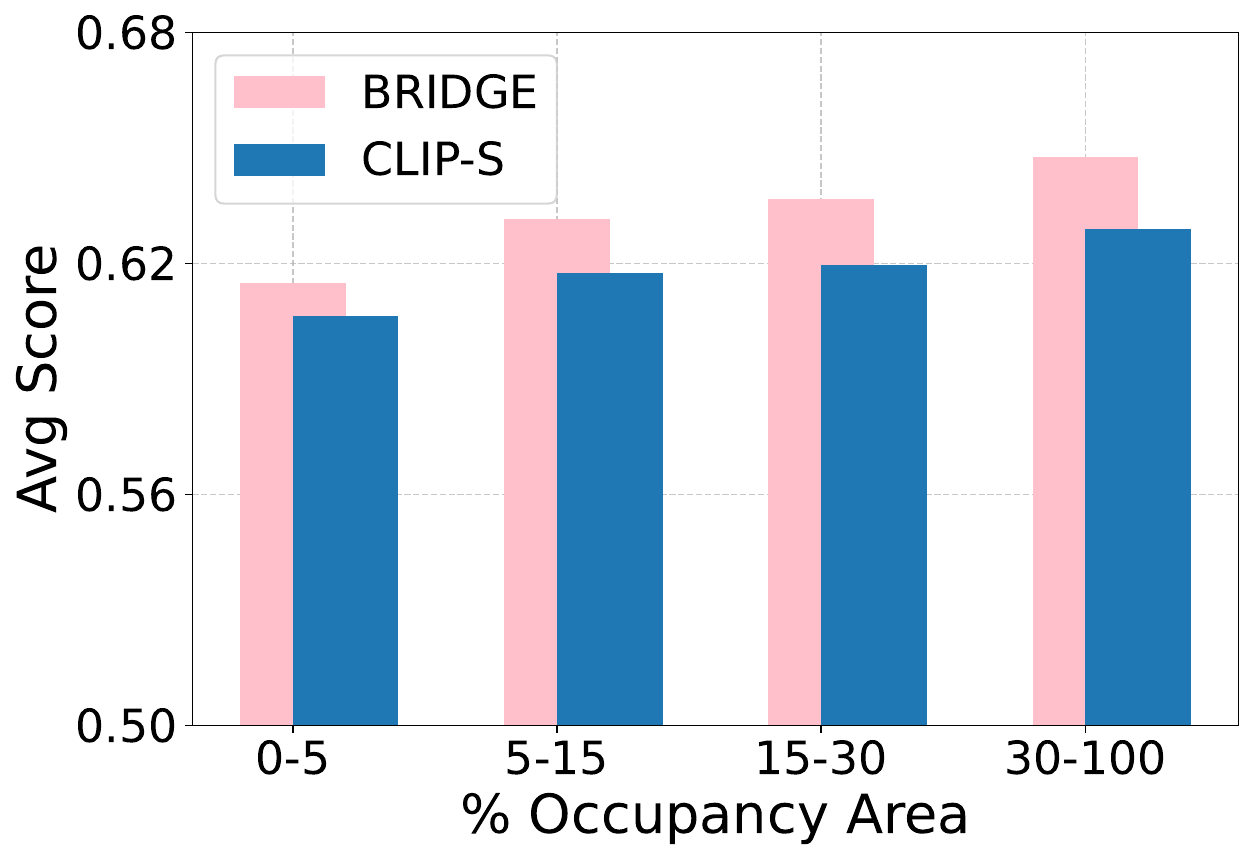} 
    \end{tabular}
}
\vspace{-0.45cm}
\caption{Metric scores for top-$k$ detections ranked by probability (left) and as a function of detection area (right).}
\label{fig:finegrained}
\vspace{-0.45cm}
\end{figure}

In the bottom part of Table~\ref{tab:ablation_supp}, we report the results ablating other architectural choices. In particular, instead of taking the output of the mapping module in the correspondence of the \texttt{[MASK]} tokens, we feed the entire output to the textual encoder. Employing this model variant, referred to as ``w/ entire mapping module output'', when computing the \ours metric leads to significantly lower correlation scores.

We also investigate the impact of using global image features instead of grid-level features. In \ours, we opt for employing grid-level features because the mapping module integrates visual features into the pseudo-caption, necessitating high-quality and detailed visual encoder outputs to effectively enhance the pseudo-captions. 
Indeed, upon examining the results, we observe a drop in the performance when using global visual features instead of more fine-grained grid features. This not only reinforces our choice but also emphasizes that our model can capture more robust fine-grained details by leveraging grid-level features. 

This observation is also supported by the analysis reported in Fig.~\ref{fig:finegrained}.
In particular, we extract object detections using the Grounding DINO model~\cite{liu2023grounding} and compute scores between the image and the prompted class name (``\texttt{a photo of a $<$class$>$}'') of each detected object. To ensure a fair comparison, we adjust \ours distribution to match that of CLIP-S. As it can be seen, \ours demonstrates higher confidence for larger objects. However, compared to CLIP, it assigns higher scores to all detections, indicating greater confidence even in smaller objects. Moreover, in the FOIL dataset, where a single detail is changed in the caption, \ours demonstrates its stronger fine-grained capability by identifying hallucinated objects better than CLIP-S and PAC-S (cf. Table~\ref{tab:foil} of the main paper).

\begin{figure*}[t]
\centering
\includegraphics[width=0.95\linewidth]{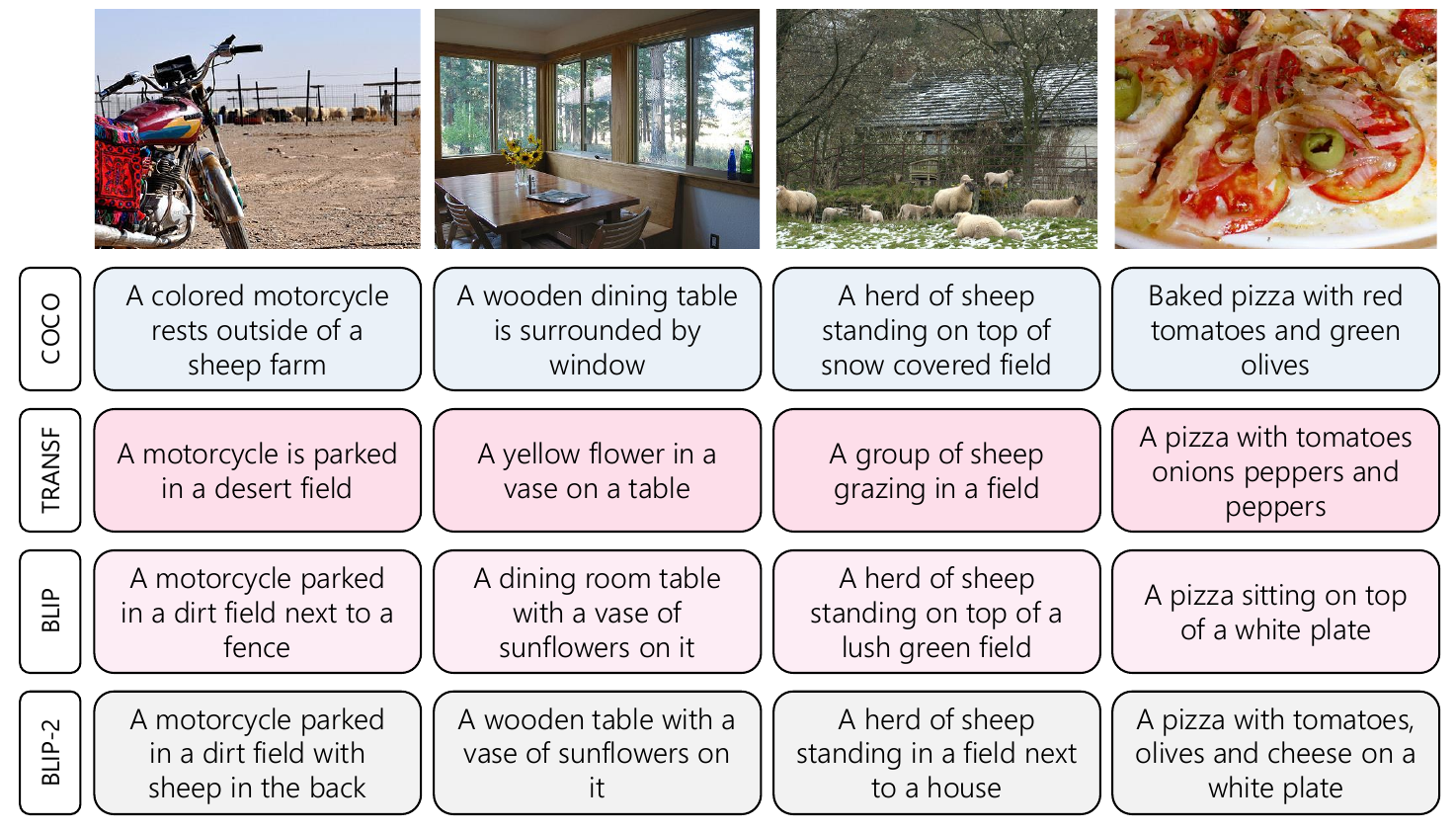}
\vspace{-0.2cm}
\caption{Sample captions generated by a standard Transformer model, BLIP, and BLIP-2, in comparison with ground-truth textual sentences from the COCO test set.}
\label{fig:generated_supp}
\vspace{-0.4cm}
\end{figure*}

\tit{Qualitative Results}
To qualitatively validate generated templates, we report in Fig.~\ref{fig:generated_supp} additional sample captions generated by the three considered models in comparison to a ground-truth caption from the COCO test set. In particular, captions generated by BLIP-2 are generally more detailed and better describe the visual content of the input image compared to those generated by BLIP and, especially, the standard Transformer model.

\begin{figure*}[!]
\centering
\includegraphics[width=0.95\linewidth]{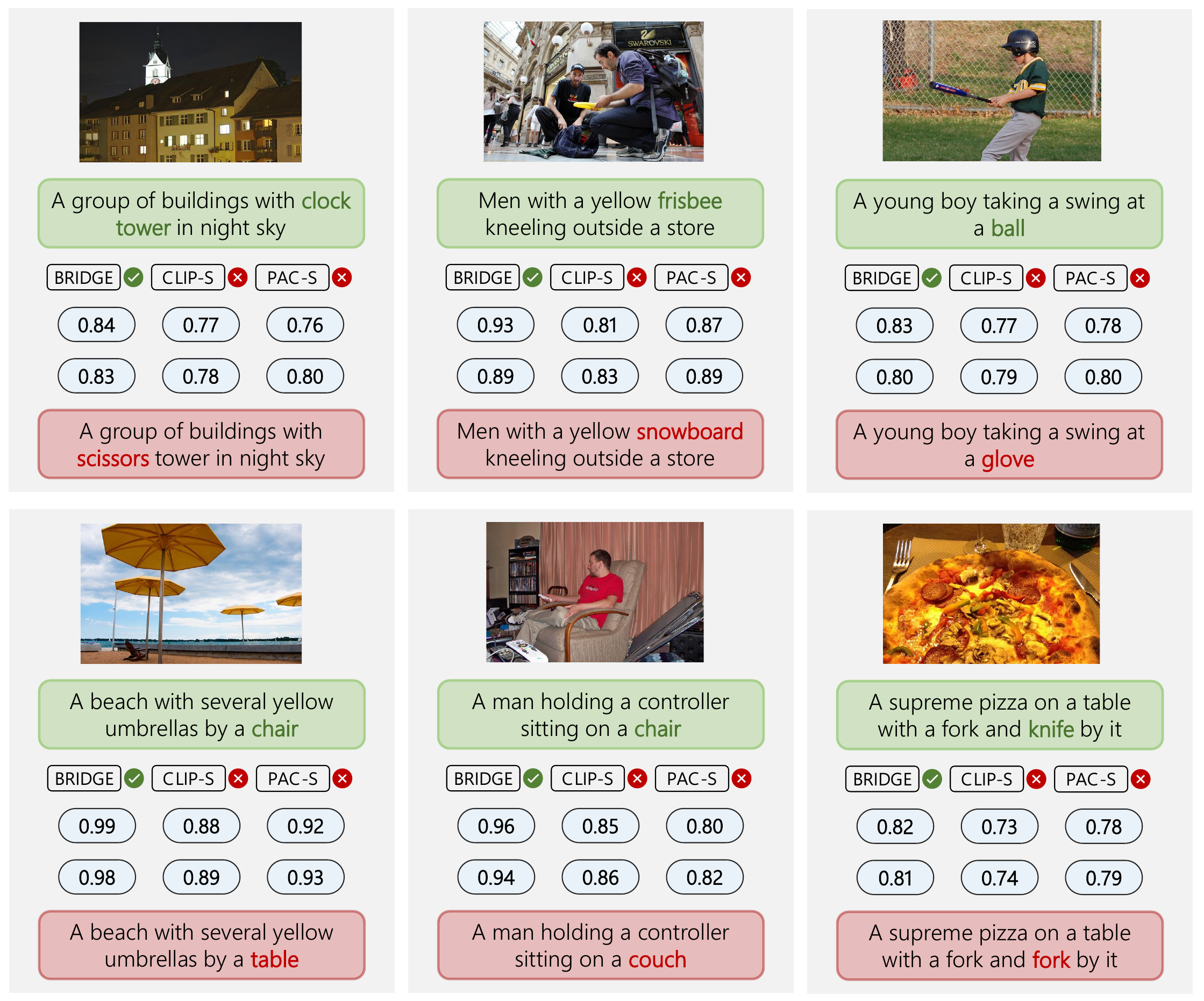}
\vspace{-0.2cm}
\caption{Sample images from the FOIL hallucination detection dataset and corresponding evaluation scores generated by the \ours metric in comparison with CLIP-S and PAC-S. Hallucinated objects are highlighted in red.}
\label{fig:foil_supp}
\end{figure*}

\begin{figure*}[t]
\centering
\includegraphics[width=0.95\linewidth]{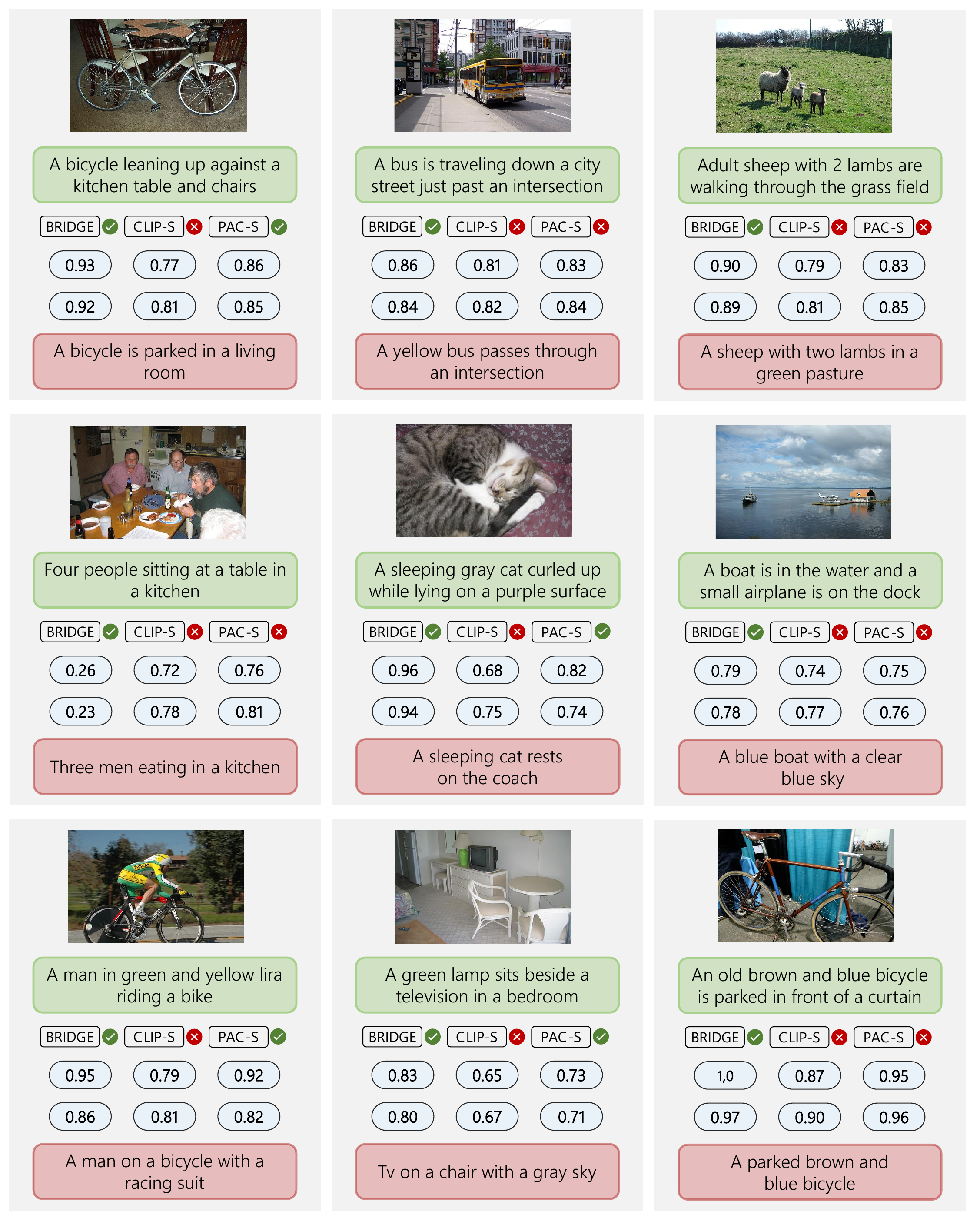}
\vspace{-0.2cm}
\caption{Comparisons of recent metrics for captioning with respect to  \ours on the Pascal-50S dataset. The candidate caption in green is the one preferred by humans.}
\label{fig:pascal_supp}
\end{figure*}

In Fig.~\ref{fig:foil_supp}, we present qualitative results on the FOIL dataset. We report results comparing \ours to CLIP-S and PAC-S, showing that our proposed metric achieves better results in terms of detection of hallucinated objects. Additional comparisons of our metric with CLIP-S and PAC-s on the Pascal50-S dataset are presented in Fig.~\ref{fig:pascal_supp}. Observing the results, although PAC-S, in some instances, aligns with human judgment, CLIP-S consistently assigns a lower score to the caption preferred by humans. On the other hand, \ours metric demonstrates its effectiveness across the majority of cases.

\end{document}